\documentclass{article}

\usepackage[preprint]{neurips_2024}

\usepackage[utf8]{inputenc} 
\usepackage[T1]{fontenc}    
\usepackage{url}            
\usepackage{booktabs}       
\usepackage{amsfonts}       
\usepackage{nicefrac}       
\usepackage{microtype}      
\usepackage{xcolor}         
\usepackage{adjustbox}

\usepackage{multirow}
\usepackage{float}
\usepackage[utf8]{inputenc} 
\usepackage[T1]{fontenc}    
\usepackage{url}            
\usepackage{booktabs}       
\usepackage{amsfonts}       
\usepackage{nicefrac}       
\usepackage{microtype}      
\usepackage{mathtools}
\usepackage{natbib}
\usepackage{xcolor}
\usepackage{url}
\usepackage{amsmath}
\usepackage{amssymb}
\usepackage{amsthm}
\usepackage{graphicx}
\usepackage{nicefrac}
\usepackage{appendix}
\usepackage{enumitem}
\usepackage{xspace}
\usepackage{pifont}
\usepackage{tikz}

\usetikzlibrary{positioning,fit,backgrounds,arrows.meta,shapes.geometric,calc,shapes.symbols}

\usepackage{wrapfig}
\usepackage{caption}
\usepackage{subcaption}
\usepackage{array}
\definecolor{darker}{rgb}{0,0.15,0.65}
\usepackage[colorlinks, urlcolor=darker, citecolor=darker, linkcolor=darker]{hyperref} 

\usepackage{colortbl}
\definecolor{green}{HTML}{009B55}

\usepackage[utf8]{inputenc} 
\usepackage[T1]{fontenc}    
\usepackage{hyperref}       
\usepackage{url}            
\usepackage{booktabs}       
\usepackage{amsmath}        
\usepackage{tikz}           
\usepackage{siunitx}        
\usepackage{amssymb}        
\usepackage{amsfonts}       
\usepackage{nicefrac}       
\usepackage{microtype}      
\usepackage{xcolor}         
\usepackage{graphicx}       
\usepackage{multirow}
\usepackage{svg} 

\usetikzlibrary{arrows.meta, positioning, fit, calc, shapes.geometric, shadows, backgrounds,shapes.symbols}

\newif\ifshowtodos
\showtodostrue   

\title{EntWorld: A Holistic Environment and Benchmark for Verifiable Enterprise GUI Agents}

\author{%
  Ying Mo\textsuperscript{\rm 1}   
  Yu Bai\textsuperscript{\rm 1}   
  Dapeng Sun\textsuperscript{\rm 1} 
  Yuqian Shi\textsuperscript{\rm 1}   
  Yukai Miao\textsuperscript{\rm 1}  
  Li Chen\textsuperscript{\rm 1}  
  Dan Li\textsuperscript{\rm 2} \\
 \textsuperscript{\rm 1}Zhongguancun Laboratory   \textsuperscript{\rm 2}Tsinghua University \\
  \texttt{\{moy, baiyu, sundp, shiyuqian, miaoyk, lichen\}@zgclab.edu.cn} \\
  \texttt{tolidan@tsinghua.edu.cn} \\
}

\begin{document}

\maketitle


\begin{abstract}
Recent advances in Multimodal Large Language Models (MLLMs) have enabled agents to operate in open-ended web and operating system environments. 
However, existing benchmarks predominantly target consumer-oriented scenarios (e.g., e-commerce and travel booking), failing to capture the complexity and rigor of professional enterprise workflows. 
Enterprise systems pose distinct challenges, including high-density user interfaces, strict business logic constraints, and a strong reliance on precise, state-consistent information retrieval-settings in which current generalist agents often struggle.
To address this gap, we introduce EntWorld, a large-scale benchmark consisting of 1,756 tasks across six representative enterprise domains, including customer relationship management (CRM), information technology infrastructure library (ITIL), and enterprise resource planning (ERP) systems. 
Unlike previous datasets that depend on fragile execution traces or extensive manual annotation, EntWorld adopts a schema-grounded task generation framework that directly reverse-engineers business logic from underlying database schemas, enabling the synthesis of realistic, long-horizon workflows. 
Moreover, we propose a SQL-based deterministic verification mechanism in building datasets that replaces ambiguous visual matching with rigorous state-transition validation.
Experimental results demonstrate that state-of-the-art models (e.g., GPT-4.1) achieve 47.61\% success rate on EntWorld, substantially lower than the human performance, highlighting a pronounced enterprise gap in current agentic capabilities and the necessity of developing domain-specific agents. 
We release EntWorld as a rigorous testbed to facilitate the development and evaluation of the next generation of enterprise-ready digital agents.
\end{abstract}

\section{Introduction}
\label{sec:intro}

The rapid evolution of Multimodal Large Language Models (MLLMs) has catalyzed a paradigm shift from passive chatbots to autonomous \textbf{Computer-Using Agents (CUAs)} capable of perceiving and interacting with complex Graphical User Interfaces (GUIs) \citep{xie2024osworld, anthropic2024computeruse}. 
While recent agents have demonstrated proficiency in general web browsing \citep{zhou2023webarena, he2024webvoyager} and mobile device manipulation \citep{rawles2024androidworld, zhang2023appagent}, a critical frontier remains underexplored: \textbf{Enterprise Automation}.
In professional settings, ``Digital Employees'' are expected to navigate vertical software ecosystems (e.g., ERP, CRM, ITSM) to execute rigorous workflows—from financial reconciliation to supply chain management.

However, benchmarking agents in this enterprise domain presents unique challenges that existing datasets hard to address:

\begin{itemize}[leftmargin=*, itemsep=0.3em, topsep=0em]
    \item \textbf{Shallow Navigation vs. Deep Logic Dependence:} Most existing benchmarks focus on query-based tasks or stateless navigation (e.g., ``buy a red dress'') \citep{deng2023mind2web, yao2022webshop}. In contrast, enterprise workflows are characterized by strict \textit{state dependencies} and long-horizon logic (e.g., ``an invoice cannot be posted until the purchase order is approved and inventory is verified''). Current general-purpose OS benchmarks lack the domain-specific logic depth to stress-test these capabilities \citep{xie2024osworld}.
    
    \item \textbf{The Data Scarcity Wall \& Privacy:} Unlike public web data, authentic enterprise trajectories are protected by strict privacy regulations and require expensive domain expertise to annotate \citep{cao2024spider2v}. Approaches relying on large-scale human demonstrations \citep{mu2025gui360} face severe scalability bottlenecks when adapting to diverse enterprise software versions.
    
    \item \textbf{Evaluation Hallucination:} A growing number of benchmarks rely on ``LLM-as-a-Judge'' or visual similarity metrics to assess agent performance \citep{zheng2023judging}. As highlighted in recent safety studies \citep{kuntz2025osharm}, probabilistic judges suffer from alignment bias and cannot verify latent system states (e.g., \textit{``Did the database transaction actually commit?''}), rendering them unsuitable for high-stakes business operations where precision is paramount.
\end{itemize}

To bridge this gap, we introduce \textbf{EntWorld}, a scalable, interactive, and deterministically verifiable environment designed for the next generation of enterprise agents.
Departing from web-based datasets, EntWorld leverages a novel schema-driven task synthesis engine. 
By reverse-engineering the database schemas and business logic of open-source enterprise systems (e.g., EspoCRM), we programmatically generate valid initial states and ground-truth workflows. 
This approach allows us to construct a fully functional dockerized enterprise aandbox, creating a privacy-free testbed that mimics the complexity of real-world operations.

Our main contributions are summarized as follows:
\begin{itemize}[leftmargin=*, itemsep=0.3em, topsep=0em]
    \item \textbf{The EntWorld Environment:} We present a multi-app enterprise sandbox integrating 6 core business applications (ERP, CRM, ITSM, etc.). It supports multi-turn interaction via a standard observation space (Screenshots + Accessibility Tree), compatible with SOTA agent frameworks \citep{qin2025ui}.
    
    \item \textbf{Deterministic Evaluation Protocol:} Addressing the limitations of existing web task evaluation during dataset construction, we propose a rigorous evaluation metric based on SQL state verification. By directly querying the underlying databases of the applications, EntWorld enables precise validation of task completion (e.g., verifying exact database record insertions or updates), ensuring deterministic and noise-free evaluation. 
    This approach eliminates ambiguities in visual matching and enables high-precision correctness assessment.

    \item \textbf{The ``Enterprise Gap'' Discovery:} We benchmark LLM and VLM-based agents, including GPT-4.1, Claude 3.5 Sonnet, UI-TARS \citep{qin2025ui} and Qwen-VL series-based models.
    Extensive evaluation highlights room for improvement in their performance as effective computer assistants.
    The best model achieves a task success rate 56.89\%, which has a gap with human performance 85\% of the enterprise web tasks, primarily due to the challenges in GUI grounding and insufficient operational knowledge. The comprehensive analysis about EntWorld offers valuable insights into the shortcomings of current multimodal agents and shows that current VLMs struggle severely with the high-density UI grounding \citep{hui2025winspot} and implicit logic constraints inherent to enterprise software.
\end{itemize}

\section{Related Work}
\label{sec:related_work}

\paragraph{Web and Mobile Agent Benchmarks.}
Early research in GUI automation primarily focused on web browser environments due to their structured DOM representations. 
\textbf{MiniWoB++} \citep{shi2017world} and \textbf{WebShop} \citep{yao2022webshop} established foundational environments for instruction following in simplified, synthetic web settings. 
To address the gap with real-world complexity, \textbf{Mind2Web} \citep{deng2023mind2web} collected over 2,000 open-ended tasks from real websites, while \textbf{WebArena} \citep{zhou2023webarena} introduced a reproducible, self-hosted environment to evaluate long-horizon web tasks.
Parallel efforts in the mobile domain include \textbf{AndroidWorld} \citep{rawles2024androidworld} and \textbf{AppAgent} \citep{zhang2023appagent}, which evaluate agents on dynamic mobile operating systems.
However, these benchmarks are confined to specific sandboxes (browsers or mobile emulators) and lack the multi-application interoperability required for professional enterprise workflows.

\paragraph{General Desktop OS Environments.}
Recent work has expanded agent capabilities to full desktop operating systems. 
\textbf{OSWorld} \citep{xie2024osworld} serves as a pioneering multimodal benchmark supporting Ubuntu, Windows, and macOS, covering open-ended tasks across diverse native applications. 
Similarly, \textbf{GUI-360$^\circ$} \citep{mu2025gui360} and \textbf{OSUniverse} \citep{davydova2025osuniverse} provide comprehensive datasets for Windows visual grounding and action prediction.
While these benchmarks assess general computer use proficiency (e.g., file management, media editing), they often lack the strict \textit{business logic constraints} and \textit{interdependent state changes} characteristic of enterprise resource planning (ERP) or customer relationship management (CRM) systems.

\paragraph{Enterprise and Vertical Domain Automation.}
Automating professional workflows presents unique challenges in logic depth and verification \citep{miao2023empiricalstudynetopscapability,wang2025digitalcybersecurityexpertfar}. 
\textbf{WorkArena} \citep{drouin2024workarena} leverages the ServiceNow platform to benchmark agents on IT service management tasks. 
More recently, \textbf{Spider2-V} \citep{cao2024spider2v} introduced a rigorous benchmark for data science and engineering workflows, requiring agents to manipulate professional tools like Snowflake and BigQuery. 
Unlike \textbf{EntWorld}, which employs deterministic SQL-based state verification to ensure rigorous evaluation, many existing enterprise benchmarks rely on heuristic string matching or ``LLM-as-a-Judge,'' which can suffer from evaluation hallucinations \citep{kuntz2025osharm}. 
Our work bridges this gap by focusing on high-density enterprise UIs with deterministic execution outcomes.

\paragraph{GUI Grounding and Visual Agents.}
The efficacy of GUI agents relies heavily on visual grounding capabilities. 
Recent advancements such as \textbf{SeeClick} \citep{cheng2024seeclick} and \textbf{WinSpot} \citep{hui2025winspot} have significantly improved agents' ability to locate UI elements from raw pixels. 
Furthermore, specialized foundation models like \textbf{UI-TARS} \citep{qin2025ui} and \textbf{Ferret-UI} \citep{you2024ferret} facilitate end-to-end GUI interaction without reliance on accessibility trees. 
EntWorld integrates these advancements to stress-test agents in visually cluttered enterprise interfaces, where element density far exceeds consumer applications.


\section{EntWorld Benchmmark}
\label{sec:method}

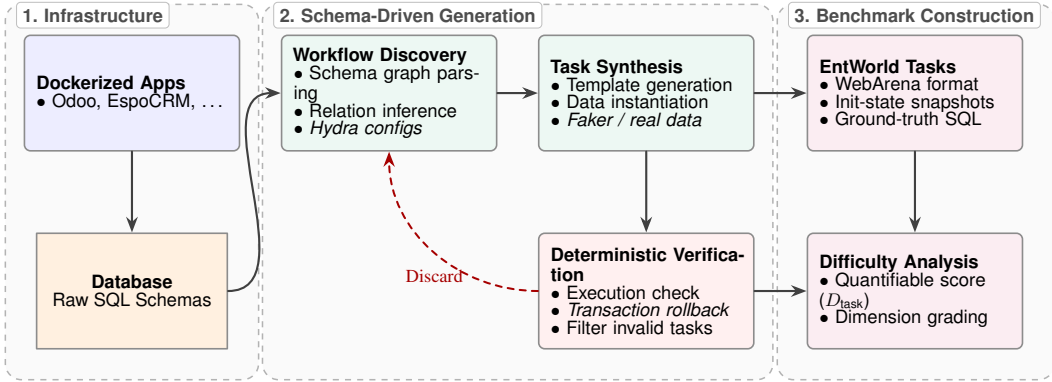
\begin{figure}[t]
\centering
\begin{adjustbox}{max width=\linewidth,center} 
\begin{tikzpicture}[
  >=Stealth,
  font=\sffamily\footnotesize,
  line cap=round, line join=round,
  node distance=15mm and 12mm,
  transform shape
]

\tikzset{box/.style={
  draw=black!45, rounded corners=3pt, line width=0.8pt,
  fill=white, inner sep=6pt,
  text width=33mm, minimum height=16mm,
  align=left
}}
\tikzset{boxBlue/.style={box, fill=blue!7}}
\tikzset{boxGreen/.style={box, fill=green!7}}
\tikzset{boxRed/.style={box, fill=red!6, text width=33mm}}
\tikzset{boxPurple/.style={box, fill=purple!7}}

\tikzset{db/.style={
  draw,
  shape=database,
  draw=black!45, fill=orange!12, shape=database,
  line width=0.8pt, inner sep=3pt,
  minimum width=33mm, minimum height=16mm, align=center
}}

\tikzset{group/.style={
  draw=black!30, dashed, rounded corners=8pt,
  line width=0.75pt, fill=black!2,
  minimum width=0mm, minimum height=55mm, 
  inner xsep=9pt, inner ysep=13pt
}}
\tikzset{grouptag/.style={
  anchor=west, font=\sffamily\bfseries\small,
  text=black!70, fill=white, draw=black!25,
  rounded corners=2pt, inner sep=3pt
}}

\tikzset{flow/.style={->, draw=black!75, line width=1.05pt}}
\tikzset{feedback/.style={->, draw=red!65!black, line width=1.0pt, dashed}}

\node[boxBlue,minimum height=20mm] (docker) {%
  \textbf{Dockerized Apps}\\[-0.15em]
  {\footnotesize $\bullet$ Odoo, EspoCRM, \ldots}
};

\node[db, shape=rectangle, below=14mm of docker,minimum height=20mm] (dbnode) {%
  \textbf{Database}\\[-0.15em]
  {\footnotesize Raw SQL Schemas}
};

\node[boxGreen, right=7mm of docker,minimum height=20mm] (discover) {%
  \textbf{Workflow Discovery}\\[-0.15em]
  {\footnotesize
    $\bullet$ Schema graph parsing\\[-0.15em]
    $\bullet$ Relation inference\\[-0.15em]
    $\bullet$ \textit{Hydra configs}
  }
};

\node[boxGreen, right=7mm of discover,minimum height=20mm] (synthesize) {%
  \textbf{Task Synthesis}\\[-0.15em]
  {\footnotesize
    $\bullet$ Template generation\\[-0.15em]
    $\bullet$ Data instantiation\\[-0.15em]
    $\bullet$ \textit{Faker / real data}
  }
};

\node[boxRed, below=14mm of synthesize, minimum height=20mm] (verify) {%
  \textbf{Deterministic Verification}\\[-0.15em]
  {\footnotesize
    $\bullet$ Execution check\\[-0.15em]
    $\bullet$ \textit{Transaction rollback}\\[-0.15em]
    $\bullet$ Filter invalid tasks
  }
};
\node[boxPurple, right=9mm of synthesize,minimum height=20mm] (bench) {%
  \textbf{EntWorld Tasks}\\[-0.15em]
  {\footnotesize
    $\bullet$ WebArena format\\[-0.15em]
    $\bullet$ Init-state snapshots\\[-0.15em]
    $\bullet$ Ground-truth SQL
  }
};

\node[boxPurple, below=14mm of bench,minimum height=20mm ] (diff) {%
  \textbf{Difficulty Analysis}\\[-0.15em]
  {\footnotesize
    $\bullet$ Quantifiable score ($D_{\text{task}}$)\\[-0.15em]
    $\bullet$ Dimension grading
  }
};

\begin{scope}[on background layer]
  \node[group, fit=(docker) (dbnode), minimum height=65mm] (g1) {};
  \node[grouptag] at ($(g1.north west)+(2mm,-2mm)$) {1. Infrastructure};

  \node[group, fit=(discover) (synthesize) (verify), minimum height=65mm] (g2) {};
  \node[grouptag] at ($(g2.north west)+(2mm,-2mm)$) {2. Schema-Driven Generation};

  \node[group, fit=(bench) (diff) , right=0.5mm of g2, minimum width=48mm, minimum height=65mm] (g3) {}; 
  \node[grouptag] at ($(g3.north west)+(2mm,-2mm)$) {3. Benchmark Construction};
  
\end{scope}

\draw[flow] (docker) -- (dbnode);
\draw[flow] (dbnode.east) to[out=0,in=180] (discover.west);

\draw[flow] (discover) -- (synthesize);
\draw[flow] (synthesize) -- (verify);

\draw[feedback] (verify.west) to[out=180,in=-90]
  node[pos=0.30, left, font=\footnotesize, text=red!65!black] {Discard}
  (discover.south);

\draw[flow] (synthesize.east) to[out=0,in=180] (bench.west);
\draw[flow] (verify.east) to[out=0,in=180] (diff.west);
\draw[flow] (bench.south) -- (diff.north);

\end{tikzpicture}
\end{adjustbox}

\caption{\textbf{Overview of the EntWorld construction pipeline.} The construction process consists of workflow discovery, verifiable task generation, benchmark construction, and quantitative difficulty analysis. Our benchmark EntWorld environment can run in parallel on a single host machine, enabling efficient learning and evaluation. }
\label{fig:method_overview}
\end{figure}

We propose a bottom-up, schema-driven framework to construct \textbf{EntWorld}. 
Unlike previous approaches that rely on manual annotation or fragile UI recording \citep{mu2025gui360}, our pipeline automatically reverse-engineers business logic from the underlying databases to synthesize verifiable tasks.
As shown in Figure \ref{fig:method_overview}, the framework consists of four pipelined modules: \textit{Workflow Discovery}, \textit{Verifiable Task Generation}, \textit{Benchmark Construction}, and \textit{Quantitative Difficulty Analysis}.

\subsection{Infrastructure \& Configuration Management}
To ensure adaptability across diverse enterprise systems (e.g., ERP and CRM), we design a layered infrastructure that promotes modularity and extensibility.

\textbf{Hierarchical Configuration.} 
Due to the diversity of target tasks and the structural complexity of business information in enterprise websites, we adopt a hierarchical inheritance and composition mechanism to reduce reliance on manual annotation. 
This mechanism modularizes various configurations, such as site-specific business information and generation parameters. 
Task generation strategies can be adjusted solely by modifying the configuration file, which significantly improves system adaptability and maintainability.

\textbf{Database Adapter Layer.} 
We abstract a \texttt{DatabaseAdapter} to encapsulate low-level access details.
Adapter instances are dynamically instantiated via factory methods according to the configuration.
This design decouples task generation logic from the underlying database engine, ensuring that higher-level components remain fully agnostic to the specific database implementation.

\subsection{Automated Workflow Discovery}
\label{subsec:workflow_discovery}
To mitigate the need for expensive domain exper annotation, we introduce a module  \texttt{HydraUniversalWorkflowDiscoverer} automating the abstraction of business workflows from raw database schemas. 
This process has the following core subsections.

\textbf{Schema Parsing \& Semantic Analysis.} 
The system first queries the database schema to filter non-empty candidate tables. It then prompts an LLM to generate a \texttt{SchemaJSON} for each table, inferring business purposes and key fields based on table nomenclature and content.

\textbf{Relation Inference \& Verification.} 
We reconstruct the entity-relationship graph through a hybrid approach:
\begin{itemize}
    \item \textit{Explicit Constraints:} We directly parse explicit foreign keys to establish high-confidence relationships.
    \item \textit{Implicit Association:} For tables lacking explicit foreign keys, an LLM infers potential links based on the semantic similarity of column names and data types.
    \item \textit{SQL Verification:} To eliminate hallucinations, the system executes probe SQL queries on inferred connections. Only relationships that return valid join results are retained and marked as high-confidence.
\end{itemize}
    
\textbf{Workflow Abstraction.} 
Aggregating the verified relationships and table semantics, the system generates a list of abstract workflows (containing core tables, key relationships, and business descriptions). 
To optimize efficiency, we implement a caching mechanism based on the MD5 hash of core table lists, allowing reuse of analysis results across runs \citep{cao2024spider2v}.

\subsection{Verifiable Task Generation}
\label{subsec:task_generation}
This module leverages workflow configurations and verified database schema relationships, together with business scenarios and real database contents, to automatically generate a set of executable GUI agent tasks equipped with deterministic verification mechanisms. 
This section supports the end-to-end pipeline, from workflow parsing to the construction of the GUI agent benchmark. 
It consists of the following components.

\textbf{Template Synthesis.} 
Constrained by the verified schema graph, an LLM generates task templates that include bilingual (English and Chinese) natural language prompts, SQL logic, and placeholder definitions.

\textbf{Data-Driven Instantiation.} 
The system executes cross-table queries to populate templates with authentic data from the database. 
For tasks involving state changes (Create/Update/Delete), we leverage LLMs to generate realistic data variants based on existing records.

\textbf{Execution-based Verification.} 
A critical innovation is our rigorous verification protocol. 
We execute the synthesized SQL queries within a transaction block to verify executability and non-empty results. For create, update, and delete (CUD) operations, we validate the affected rows prior to issuing a \textit{Transaction Rollback}. 
This ensures the seed database remains clean and consistent for subsequent generation steps. 


\begin{figure*}[t]
    \centering
    \begin{subfigure}{0.39\textwidth}
        \centering
        \includegraphics[width=1\textwidth]{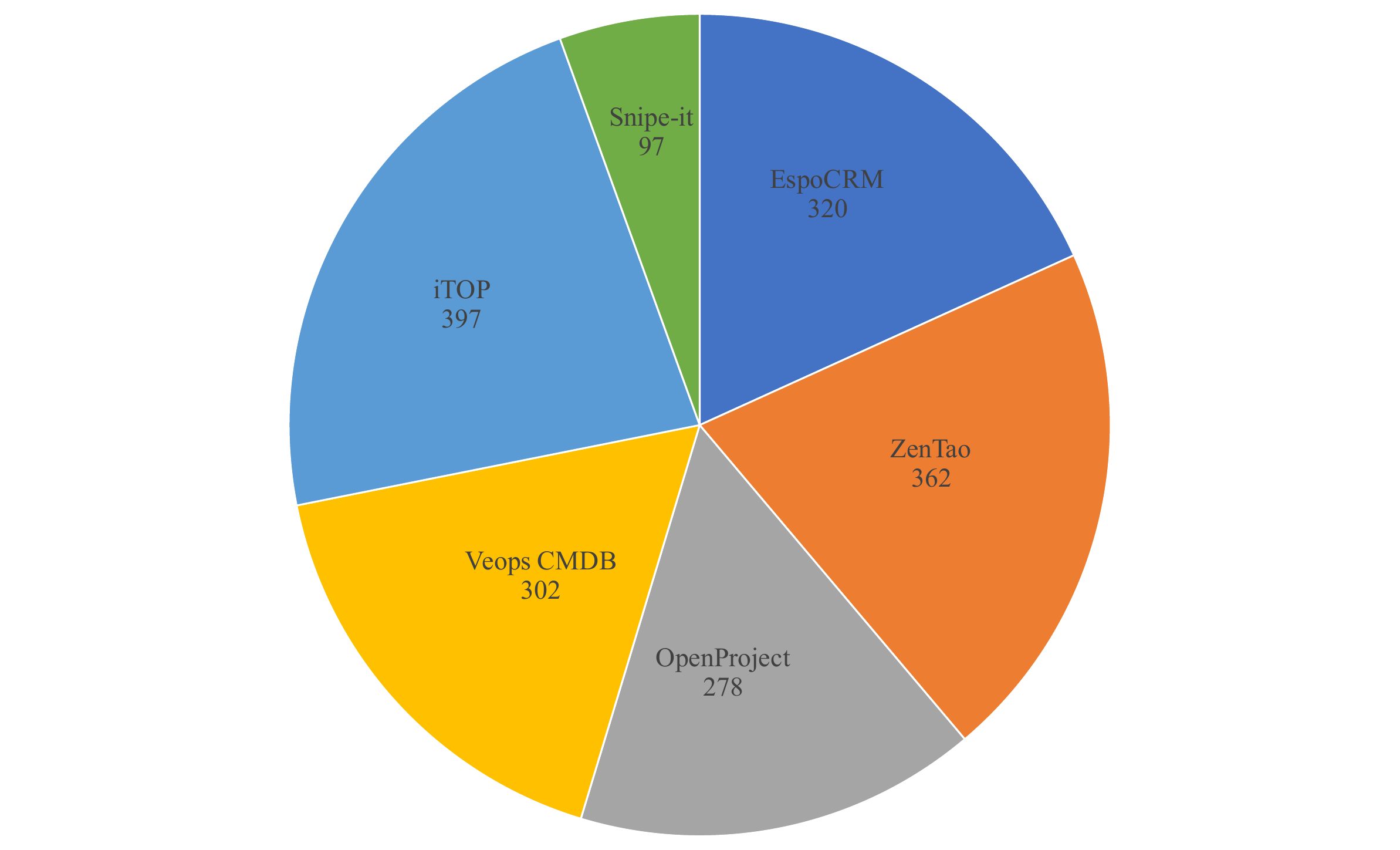}
        \caption{}
        \label{fig:benckmark_distribute}
    \end{subfigure}
    \hfill
    \begin{subfigure}{0.6\textwidth}
        \centering
        \includegraphics[width=\textwidth]{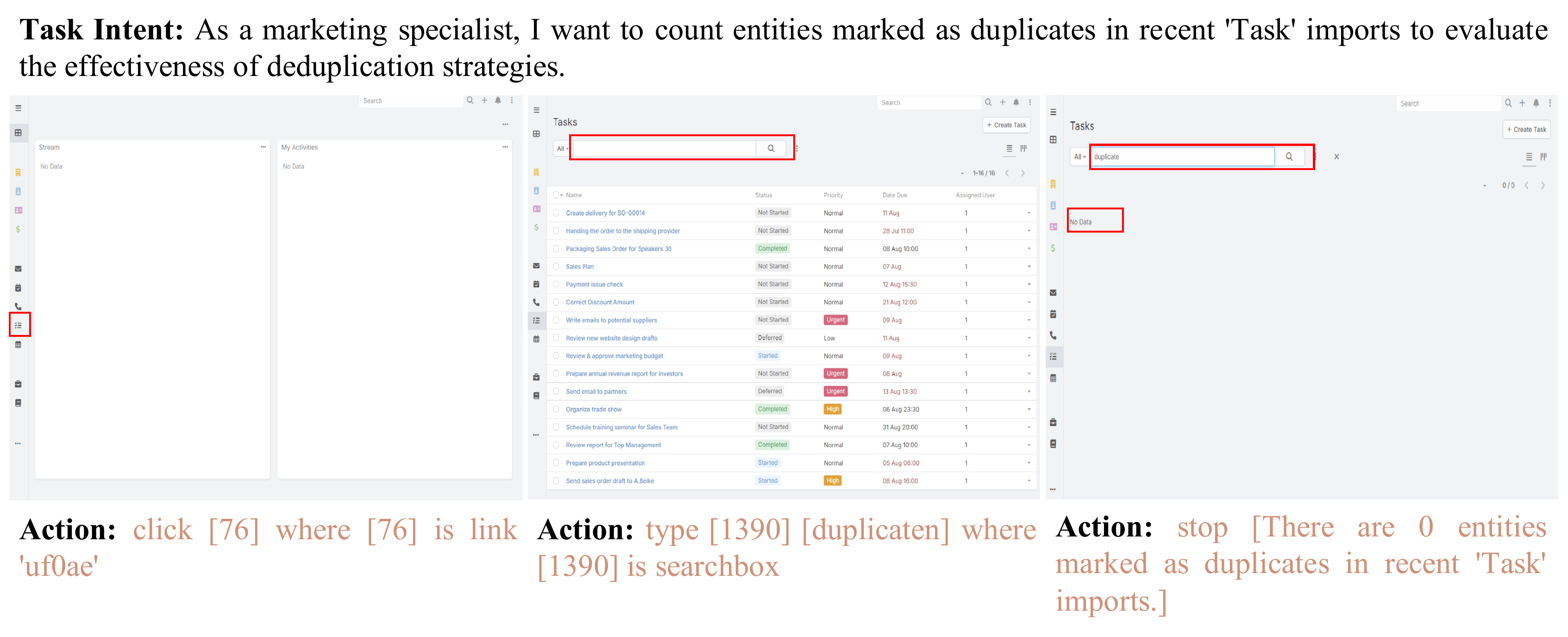}
        \caption{}
        \label{fig:benchmark_example}
    \end{subfigure}
    \caption{
        Statistics and example of the EntWorld dataset.
        (\subref{fig:benckmark_distribute}) shows the distribution of task intents in EntWorld across domians, with numbers indicating the count of sub-domains.
        (\subref{fig:benchmark_example}) illustrates a sample data instance of our dataset with the webpage screenshot and actions in accessibility tree, where actions in yellow will result in a transition to a new webpage.
    }
    \label{fig:benckmark_statistics}
\end{figure*}
\subsection{Benchmark Construction \& Standardization}
To align with the broader research community, the \texttt{task\_factory} module converts validated tasks into a standardized format compatible with WebArena and its spin-offs \citep{zhou2023webarena, koh2024visualwebarena, qi2025webrl}.
The pipeline performs deduplication based on instruction semantics and filters out tasks with illegal answers. Crucially, for non-query tasks (state-modifying), the system automatically tags them with \texttt{require\_reset}, ensuring that the evaluation environment is reset to the initial snapshot after each agent trajectory to maintain determinism \citep{xie2024osworld}.
In total, we curated 1756 instantiated intents. 
The distribution of intents is shown in Figure ~\ref{fig:benckmark_distribute}, which demonstrates a wide spectrum of user interaction purposes and a diverse range of task categories.

\subsection{Quantitative Difficulty Analysis}
\label{subsec:difficulty}


Unlike prior benchmarks that rely on subjective difficulty ratings, we introduce a quantifiable difficulty metric for enterprise web tasks. 
Specifically, we employ the \texttt{QuestionDifficultyAnalyzer} to assign each task a difficulty score based on multiple structural and operational factors. 
The score is computed as a weighted combination of several dimensions reflecting task complexity.
We provide the definitions of the difficulty function and detailed descriptions of each dimension (e.g., table, relational, operation, result, and SQL complexity) in Appendix~\ref{subapp:task_difficulty}.

The task difficulty score is computed as:
\begin{equation}
    D_{task} = \sum_{i}^{} w_i\cdot d_i
    \label{fun.benchmark_difficulty}
\end{equation}
where $d_i$ denotes the complexity factors and $w_i$ are predefined weights.
This multi-dimensional scoring allows for fine-grained analysis of agent capabilities across different task complexity levels.

\section{EntWorld Environment}
\label{sec:environment}
In this section, we will introduce the task setting of autonomous agents, the components of the EntWorld environment, and the supported action spaces and observations.

\subsection{Task Environment}
\label{subsec:task_environment}
Following \citep{yao2022webshop,zhou2023webarena,xie2024osworld,he2024webvoyager}, we formulate this task environment as $\mathcal{E} = (\mathcal{S},\mathcal{A},\mathcal{O},\mathcal{T},\mathcal{R})$ with state $\mathcal{S}$, action space $\mathcal{A}$, observation space $\mathcal{O}$, transition function $\mathcal{T}$, and reward function $\mathcal{R}$. 
Given the task intent $i$, the agent generates the action $a_t \in \mathcal{A}$ based on the intent and state at each time step $t$. 
After executing an action, the GUI agent observes a new state $s_t \in \mathcal{S}$ and observation $o_{t=1} \in \mathcal{O}$ from the environment, which is represented by the current web page's screenshot and textual content. 
Interaction with the environment terminates when the task is completed or reaches the max number of steps. 
Then the reward function assesses whether the task state satisfies the target condition, assigning success if achieved and failure otherwise.  

\subsection{Observation Space}
\label{subsec:observation_space}
The observation in EntWorld enviroment includes screenshot of the web page,  which can be parsing the accessibility (ally) tree and DOM information. 
Based on task and configuration requirements, web page observations are delivered to the intelligent agent in either text or visual modalities.

\subsection{Action Space}
\label{subsec:action_space}
We provide the action space that emulates the keyboard and mouse operations on web pages following previous works \citep{yao2022webshop, zhou2023webarena, xie2024osworld, he2024webvoyager}. 
The action space has click, type, hover, scroll, press, new\_tab, go\_back, go\_forward, goto, page\_focus, stop and none. 
Details of action space can be seen in Appendix \ref{app:details}. 
These operation commands are executed based on the corresponding screen coordinates. 
In this environment, the current web page of a task is parsed into many visible and interactive elements. These elements are generated by parsing the page's Document Object Model (DOM) or accessibility tree. 
They include a unique identifier ID, a text description, and page coordinates (typically the bounding box coordinates (x, y, width, height)). 
The agent generates an action element ID and determines the operation location on the page through element matching.

\section{Experiments}
\label{sec:experiments}
In this section, we present the implementation details of experimental settings for state-of-the-art LLM and VLM agent baselines on \textbf{EntWorld}, and report their corresponding performance results. 



\subsection{Experimental Setup}
\textbf{Baselines.} We evaluate a diverse set of agents including:
(1) \textbf{Proprietary Models:} \textsc{GPT-4.1} \citep{achiam2023gpt}, \textsc{Claude 3.5 Sonnet} (via Computer Use API) \citep{anthropic2024computeruse}.
(2) \textbf{Open-Weights Models:} Our benchmark covers open-weights multimodal agents such as \textsc{UI-TARS-72B-DPO}~\citep{qin2025ui} and \textsc{Qwen2.5-VL-32B-Instruct}~\citep{qwen2.5-vl}, and also includes \textsc{WebRL}~\citep{qi2025webrl}, a representative text-based web agent.
(3) \textbf{Our Models EntAgent:}
Based on Qwen2.5-VL-32B-Instruct, we develop EntAgent, an enterprise-oriented web agent trained with supervised fine-tuning (SFT) and Reinforcement learning (RL) \citep{paul2017rlhf,yao2023react,deepseek-r1}. 
Specifically, we consider two variants:  
(i) EntAgent-SFT, which is obtained via supervised fine-tuning; and
(ii) EntAgent-RL, which is further obtained via end-to-end reinforcement learning. 
For training EntAgent, a trajectory training set was specifically curated for using the diverse instruction of templates. 
These trajectories were divided into a trajectory fine-tuning set and a reinforcement learning training set according to website features and task complexity. 
For details on the training of EntAgent-RL, please refer to the Appendix \ref{app:mehods_details}.
The trajectory data were obtained through agent interactions based on different template instructions, and ultimately, trajectories that successfully completed the tasks were selected. 
(4) \textbf{Human Reference:} Following \citet{xie2024osworld}, we engaged expert annotators to establish a human performance ceiling. 

\textbf{Metrics.} 
For each task, we select appropriate evaluation functions and parameters to construct a configuration file. 
Each sample is associated with a dedicated, executable evaluation script. 
The evaluation functions determine task success by assessing extracted key components. 
Accordingly, in this experimental section, we use task \textit{success rate (SR)} as the evaluation metric. 


\subsection{Main Results}
Table \ref{tab:main_results} presents the results derived from web agents. 
GPT-4.1 achieves a task success rate of 47.61\%, which is lower than the human performance of 37.39\%. 
The gap indicates that proprietary models remain significantly inferior to human evaluators in performance. 
This result also suggests that even strong proprietary models still face notable challenges in complex, interactive enterprise environments. 
Among open-source web agents, performance is correlated with the choice of the base model. 
For instance, WebRL-9B (based on GLM-4-9B \citep{glm2024chatglm}) outperformed WebRL-8B (based on Llama-3.1-8B \citep{grattafiori2024llama3}) by a notable 15.83 percentage points. 
Furthermore, open-source multimodal agents have demonstrated competitive potential in enterprise tasks, with UI-TARS-72B-DPO achieving 34.51
Most notably, our proposed agents, EntAgent-RL and EntAgent-SFT, establish new state-of-the-art results among open-source methods, achieving success rates of 56.89\% and 50.57\%, respectively, surpassing even the leading proprietary model (GPT-4.1) in this evaluation. 
Moreover, the reinforcement‑learning‑based agent (EntAgent‑RL) outperforms its supervised fine‑tuning counterpart (EntAgent‑SFT), highlighting the advantage of RL‑based optimization in this domain. 

\begin{table*}[t]
\centering
\caption{\textbf{Main Results on EntWorld.} We report the task Success Rate (SR \%)  LLM and VLM agents across 1,756 tasks. \textbf{EntAgent-RL} (Ours) achieves a new state-of-the-art, significantly outperforming the previous best open-weights agent (UI-TARS) by \textbf{+22.38\%} and surpassing proprietary models like GPT-4.1. The cost per task is calculated as the statistical mean. }
\label{tab:main_results}
\resizebox{\textwidth}{!}{
    \begin{tabular}{l|c|c|c}
        \toprule
        \multirow{2}{*}{\textbf{Model}} & \multirow{2}{*}{\textbf{Backbone}} & \textbf{EntWorld (Overall)} & \textbf{Efficiency} \\
         & & \textbf{Success Rate (\%)} & \textbf{Cost / Task} \\
        \midrule
        \multicolumn{4}{c}{\textit{Human Baseline}} \\
        \rowcolor{gray!10} Human Performance & - & \textit{85.00}$^\dagger$   & - \\
        \midrule
        \multicolumn{4}{c}{\textit{Proprietary Agents}} \\
        GPT-4.1                                               & - & 47.61     & \$0.16 \\
        Claude 3.5 Sonnet                                     & - & 28.78$^*$ & \$0.28 \\
        \midrule
        \multicolumn{4}{c}{\textit{Open-Weights Agents}} \\
        WebRL-8B & Llama-3.1-8B                               & 2.51  & - \\
        WebRL-9B & GLM-4-9B                                   & 18.34 & -  \\
        UI-TARS-72B-DPO & Qwen2-VL-72B                        & 34.51 & - \\
        Qwen2.5-VL-32B-Instruct & Qwen2.5-VL-32B-Instruct     & 23.97 & - \\
        EntAgent-SFT (Ours) & Qwen2.5-VL-32B-Instruct         & 50.57 & - \\
        \textbf{EntAgent-RL (Ours)} & Qwen2.5-VL-32B-Instruct & \textbf{56.89}  & - \\
        \bottomrule
    \end{tabular}
}
\footnotesize{$^\dagger$Estimated human performance. $^*$Partial run on subset.}
\end{table*}


\subsection{Diagnostic Analysis}
\label{subsec:analysis}
The enterprise-scale dataset covers six representative enterprise systems, including customer relationship management (EspoCRM), project collaboration (ZenTao and OpenProject), IT asset management (Veops CMDB and Snipe-IT), and IT service management (iTOP). 
To better understand the agents’ performance across diverse environments, we conduct fine-grained analyses over operational tasks in different business domains and across varying task horizons. 

\begin{figure}[ht]
    \centering
    \begin{minipage}{0.48\textwidth}
        \centering
        \includegraphics[width=0.93\linewidth]{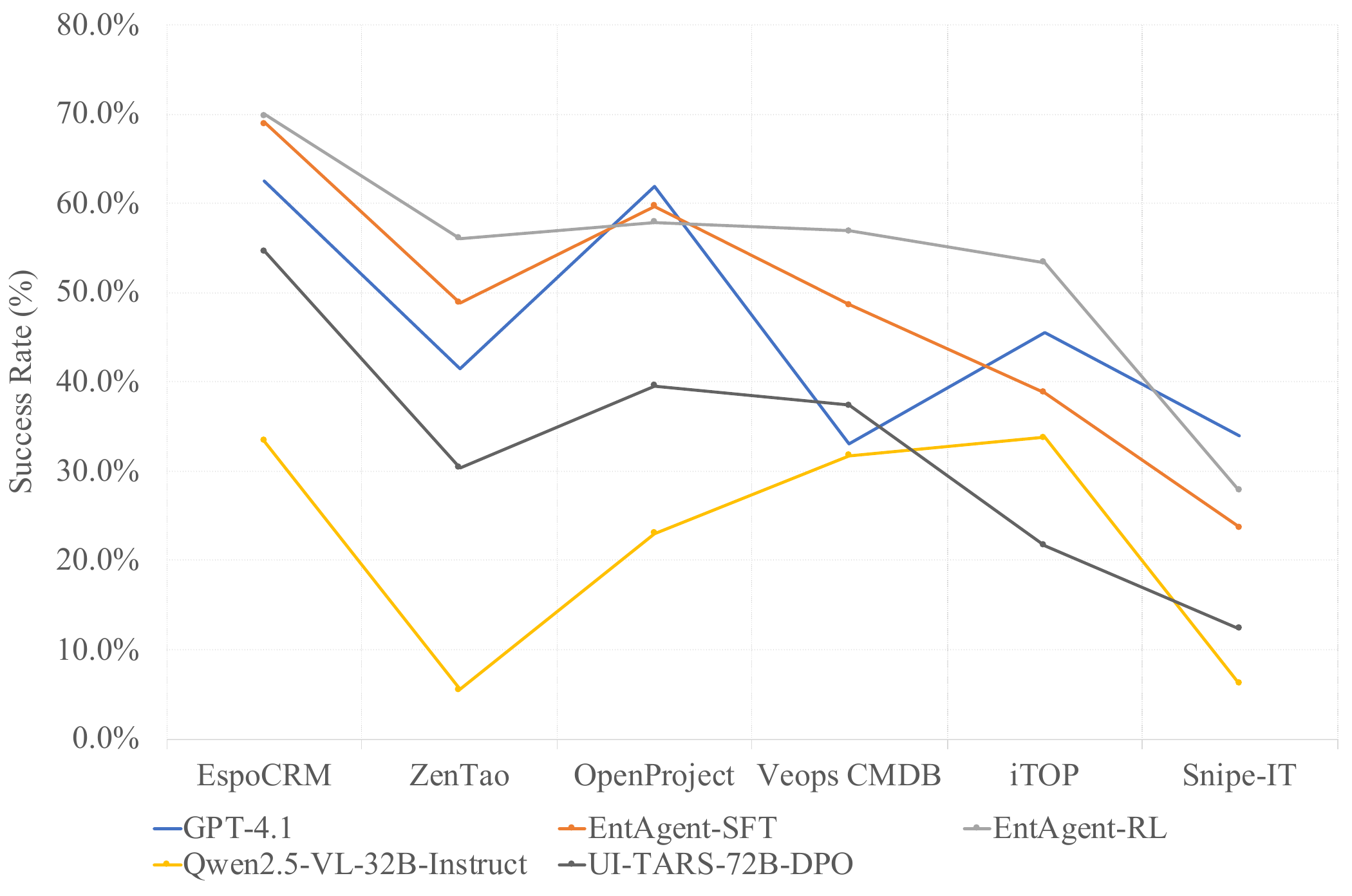}
        \caption{\textbf{Environment-Specific Analysis.} Success Rate (\%) across six distinct enterprise environments.}      
        \label{fig:benchmark_analysis_perwebsite}
    \end{minipage}
    \hfill
    \begin{minipage}{0.5\textwidth}
        \centering
        \captionof{table}{\textbf{Impact of Task Horizon.} Success Rate (\%) breakdown by the number of steps required to complete the task. Performance degrades sharply for long-horizon tasks ($>15$ steps). We evaluate performance under the screenshot and ally tree.}
        \label{tab:horizon}
        \resizebox{\columnwidth}{!}{
            \begin{tabular}{l|ccc}
                \toprule
                \textbf{Model} & \textbf{Short} ($<5$) & \textbf{Medium} ($5\text{-}15$) & \textbf{Long} ($>15$) \\
                \midrule
                GPT-4.1 & 34.0 & 13.1 & 2.7 \\
                UI-TARS-72B-DPO & 9.5 & 14.1 & 10.9 \\
                EntAgent-SFT & 8.8 & 16.2 & 25.6 \\
                EntAgent-RL & 17.4 & 16.0 & 23.5 \\
                \bottomrule
            \end{tabular}
        }
    \end{minipage}
\end{figure}

As illustrated in Figure \ref{fig:benchmark_analysis_perwebsite}, agent performance varies consistently across enterprise systems, revealing a strong correlation between success rates and task structural characteristics. 
EntAgent-RL achieves the highest or near-highest performance across most systems, particularly in environments with interaction horizons and higher state dependency.
In structured systems such as EspoCRM which focus on evaluating web tasks under clear processes and standard components, multiple agents achieve competitive performance, suggesting that deterministic workflows and explicit semantics might reduce the difficulty of end-to-end tasks. 
In contrast, for OpenProject tasks, the performance gap between reinforcement learning–based agent, supervised agent and GPT-4.1 is narrow. 
For iTOP and Snipe‑IT tasks, which exhibit both high specialization and substantial interface heterogeneity, all evaluated agents show significant performance degradation. 
This indicates that current methods still face systematic bottlenecks due to their heavy reliance on domain‑specific knowledge and limited ability to interpret non‑standard UI components.
Table \ref{tab:horizon} indicates the performance of different models under the combined input of screenshots and the accessibility tree across tasks grouped by step length, which is a key structural feature of the EntWorld benchmark, which is characterized by realistic, multi-step enterprise workflows. 
The results reveal that task horizon does not uniformly dictate model performance; instead, different models show distinct adaptation patterns to the varied procedural complexity present in the dataset.
Specifically, GPT‑4.1 excels in short-horizon tasks ($<5$ steps), achieving 34.0\% success, but its performance declines markedly in medium- and long-horizon scenarios, dropping to only 2.7\% on tasks exceeding 15 steps. 
This reflects a notable limitation in handling the extended reasoning and state-tracking demands inherent in EntWorld’s more complex workflows.
In contrast, open-source models demonstrate stronger scalability as task horizon increases. 
For instance, EntAgent‑RL maintains consistent performance across short, medium, and long-horizon tasks, illustrating its robustness in managing the sequential decision-making required by EntWorld’s elaborate procedural structures. 
Similarly, UI‑TARS‑72B‑DPO shows a progressive improvement in success rate with longer interaction sequences, while having a lower performance at long level steps (>15 steps) than the medium level, indicating that it might have limitations in strategic planning over long-horizon tasks.
These trends underscore that model performance is closely tied to how well architectures align with the benchmark’s inherent characteristics—particularly its emphasis on multi-stage, goal-oriented interactions. 
Future methods may benefit from mechanisms such as explicit state tracking, sub-goal decomposition, or episodic memory to better address the long-horizon challenges posed by datasets like EntWorld.

\begin{table}[ht]
\centering
\caption{Ablation on Input Modalities. Enterprise apps often lack clean A11y trees, making visual understanding critical.}
\label{tab:ablation}
\begin{adjustbox}{width=1\columnwidth,center}
\begin{tabular}{llccccccc}
\toprule
\multirow{2}{*}{Inputs}& \multirow{2}{*}{Model}  & \multicolumn{7}{c}{Success Rate (\%)} \\ \cline{3-9}
                       &                         & EspoCRM           & ZenTao & OpenProject & Veops CMDB & iTOP   & Snipe-IT & Avg    \\ \hline
A11y tree              & Qwen2.5-VL-32B-Instruct & 37.5\%             & 4.7\%  & 49.3\%       & 22.2\%      & 31.0\%  & 18.6\%    & 27.2\%  \\
                       & EntAgent-SFT            & 50.6\%             & 14.1\%  & 48.6\%       & 30.8\%      & 34.3\%  & 20.6\%    & 33.2\%  \\
                       & EntAgent-RL             & 53.7\%             & 14.4\%  & 49.6\%       & 35.1\%      & 34.8\%  & 23.7\%    & 35.2\%  \\ \hline
Screenshot             & Qwen2.5-VL-32B-Instruct & 13.8\%             & 7.7\%  & 15.2\%       & 9.0\%      & 11.1\%  & 7.2\%    & 10.7\%  \\
                       & EntAgent-SFT            & 5.1\%             & 22.9\%  & 19.4\%       & 27.0\%      & 10.1\%  & 7.2\%    & 15.3\%  \\
                       & EntAgent-RL             & 5.2\%             & 19.3\%  & 21.6\%       & 28.3\%      & 11.1\%  & 8.3\%    & 15.6\%  \\ \hline
Screenshot + A11y tree & Qwen2.5-VL-32B-Instruct & 33.4\%            & 5.5\%  & 23.0\%      & 31.8\%     & 33.8\%  & 6.2\%    & 22.3\% \\
                       & EntAgent-SFT            & 69.1\%            & 48.9\% & 59.7\%      & 48.7\%     & 38.8\% & 23.7\%   & 48.2\% \\
                       & EntAgent-RL             & 70.0\%            & 56.1\% & 57.9\%      & 57.0\%     & 53.4\% & 27.8\%   & 53.7\% \\
\bottomrule
\end{tabular}
\end{adjustbox}
\end{table}
\subsection{Ablation Study on Input Modality}
We investigate the impact of different observation spaces on agent performance, as summarized in Table \ref{tab:ablation}. 
Consistent with \citet{xie2024osworld}, combining screenshots with the accessibility tree (A11y tree) yields the strongest results. 
However, unlike general web agents, our evaluation reveals that relying solely on A11y tree is particularly inadequate for legacy enterprise applications, where accessibility metadata is often incomplete or inconsistently implemented.
Using only screenshots as input yields the lowest performance overall, indicating that visual input cannot provide sufficient information, or the model's understanding of visual content still contains biases. 
Pure visual GUI agents have significant room for improvement and represent a future trend. 
The performances of baseline agents only using the Ally tree as the input have better than the screenshot input in total. 
However, the significant volume of tokens in the A11y tree where even leaf nodes alone can contain tens of thousands of tokens—also introduces considerable inference overhead for the model.
The combination of Screenshot and A11y tree inputs substantially improves model performance. 
Among them, EntAgent‑RL achieves an average success rate of 53.7\% and EntAgent‑SFT reaches 48.2\%, which are better the base model Qwen2.5‑VL‑32B‑Instruct. 
These results demonstrate that multimodal information is critical for robust task understanding and reliable execution.
The performance of a given model varies significantly across different types of GUI tasks, primarily due to the diverse characteristics inherent in enterprise‑level web tasks. 
These tasks differ substantially in terms of business processes, often involving distinct multi-step workflows and structured data manipulation patterns.  
For example, a task may require sequential navigation through multiple interface layers to update a customer record, while another might demand precise form‑filling across several tabs to configure an IT asset. 
Such task‑specific heterogeneity leads to uneven performance distribution. 
These observations suggest that current methods still require further improvement in cross‑task environmental comprehension and generalizable reasoning capabilities.

\subsection{Case Studies}
\begin{figure}[h]
    \centering
    \includegraphics[width=1\textwidth]{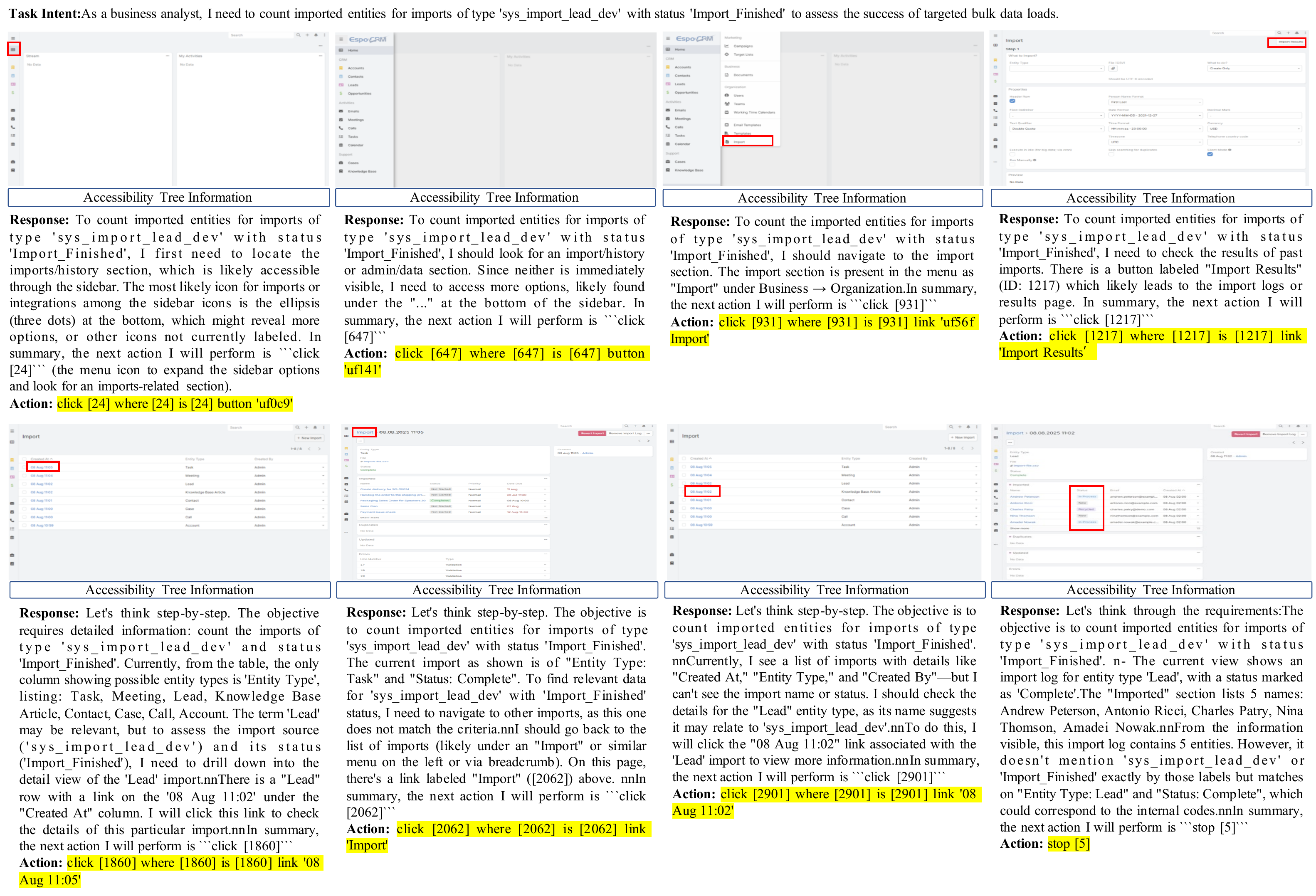}
    \caption{A case of agent failure, along with their screenshot and the accessibility tree of the relevant sections. The agent correctly interpreted the task instructions, but ultimately produced an incorrect answer due to inaccurate perception of page state changes and interface prompts. The case is about EspoCRM.}
    \label{fig:benchmark_casestudy_1}        
\end{figure}

\begin{figure}
    \centering
    \includegraphics[width=0.9\textwidth]{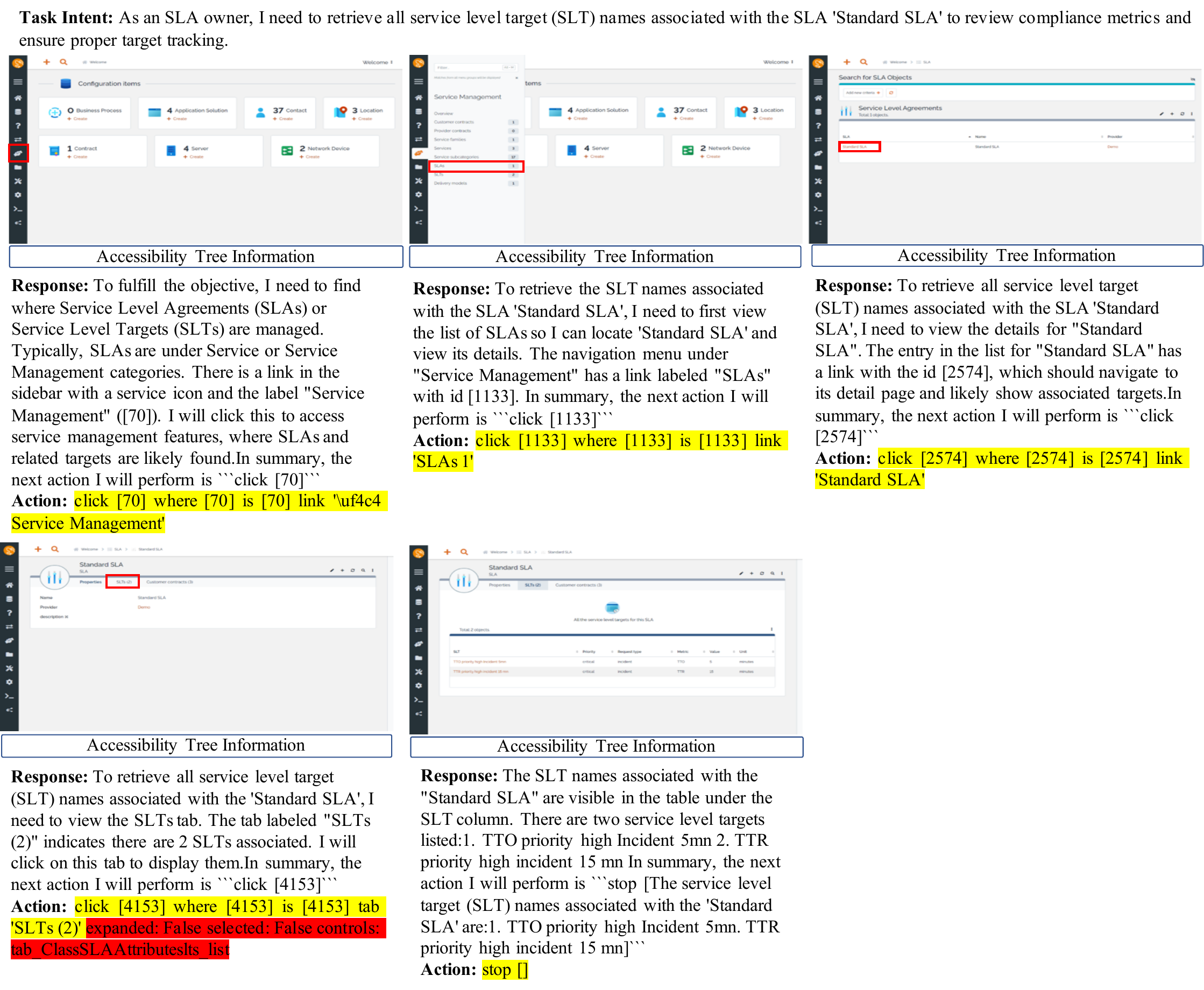}
    \caption{A case of agent failure due to UI element type misidentification, where the agent incorrectly interprets interface components and performs actions inconsistent with the actual environmental state, ultimately leading to deviation from the intended workflow. The case is about iTOP.}
    \label{fig:benchmark_casestudy_2}
\end{figure}

We present several case studies analyzing agent behaviors on end-to-end enterprise web tasks. 
GUI agents often exhibit early-stage errors during multi-step interactions, such as selecting incorrect UI elements or misinterpreting intermediate states, which subsequently lead to task failure due to error accumulation. 
As illustrated in Figure \ref{fig:benchmark_casestudy_1}, during a multi-step task, after the 5th interaction, the agent could have completed the task based on the environment observations.  
However, due to a misperception of the page content, the agent continued to execute actions. 
Even after the final operation, it still overlooked the key information displayed on the page, ultimately providing an incorrect answer.
In some failure cases, as Figure \ref{fig:benchmark_casestudy_2} shown, the agent produces actions that are inconsistent with the environment state, causing the interaction to deviate from the intent workflow. 
In some failure cases, as shown in Figure \ref{fig:benchmark_casestudy_2}, the agent misidentifies the element types (action in red), leading to actions that are inconsistent with the environmental state and causing the interaction to deviate from the intended workflow.
More cases and detailed trajectories are provided in Appendix \ref{app:case_study}.
\section{Conclusion}
\label{sec:conclusion}
We propose EntWorld, a scalable, interactive, and deterministically verifiable
environment designed for the next generation of enterprise agents. 
We present an evaluation of LLM- and VLM-based agents on six enterprise systems, and the best model achieves 56.89\% overall task success rate across 1,756 tasks. 
Key findings include: (1) significant variation in performance across system types (36--67\%), (2) UI navigation dominates the action space with 74.5\% of actions being clicks, and (3) task length (6.3--11.2 rounds average) varies significantly by system but does not directly correlate with difficulty. Analysis of complete execution traces (11,572 total actions) reveals that navigation through complex UI hierarchies is the primary challenge. Future work should extend this benchmark to additional models, include write operations beyond SELECT queries, and develop better strategies for UI element identification and navigation.




\bibliographystyle{unsrtnat} 
\bibliography{ref}


\appendix
\section{Additional Details}
\label{app:details}

\subsection{Task Distribution}
All 1,756 evaluation tasks are mainly web selection operations, focusing exclusively on information retrieval. 
The tasks span across 6 enterprise systems.
The categories by system are:

\begin{itemize}
\item Snipe-IT: 97 tasks, which is an open source asset management in IT Operations \footnote{https://www.zentao.net/}. 
\item Veops CMDB: 302 tasks,  which is a project for managing IT infrastructure and services \footnote{https://www.baidu.com}. 
\item EspoCRM: 315 tasks, which is a web management that allows users to view, input, and evaluate all your company relationships \footnote{https://www.espocrm.com/zh/}. 
\item iTOP: 397 tasks, which is web-based IT service management platform, including a fully customizable CMDB, a helpdesk system, and a document management tool\footnote{https://combodo.com/}.
\item OpenProject: 278 tasks, which is a web-based project management tool helping to manage projects, tasks and goals \footnote{https://www.openproject.org/}. 
\item ZenTao: 362 tasks, which is a project management software that covers the main PM process, from product and project management to quality management, documentation management, organization management, and office management \footnote{https://www.zentao.net/}. 
\end{itemize}


\subsection{Action Type Distribution}

Analysis of complete execution traces across all 1,756 tasks reveals the following action distribution (Table~\ref{tab:action_dist}):

\begin{table}[ht]
\caption{Action Type Distribution Across All Tasks (11,572 total actions)}
\centering
\begin{tabular}{@{}llrr@{}}
\toprule
Action Type     &Description & Count & Percentage \\
\midrule
\texttt{click}  & click an element within a webpage     & 8,619 & 74.5\% \\
\texttt{stop}   & the end of a task    & 1,315 & 11.4\% \\
\texttt{type}   & select a text box and then type new content in a webpage   & 918   & 7.9\% \\
\texttt{none}   & no operation for the webpage    & 341   & 2.9\% \\
\texttt{scroll} & the vertical movement of the entire page   & 207   & 1.8\% \\
\texttt{hover}  & hover on an element within a webpage    & 110   & 1.0\% \\
\texttt{go\_back} & return the previous webpage   & 28    & 0.2\% \\
\texttt{goto}  & go to a webpage URL     & 16    & 0.1\% \\
\texttt{press} & press a key comb     & 15    & 0.1\% \\
\texttt{new\_tab} & open a new webpage   & 2     & 0.0\% \\
\texttt{page\_focus} & focus an element within a webpage & 1    & 0.0\% \\
\bottomrule
\end{tabular}
\label{tab:action_dist}
\end{table}

The dominance of \texttt{click} actions (74.5\%) indicates that navigating through UI elements is the primary challenge in these enterprise systems. The relatively low frequency of \texttt{type} actions (7.9\%) reflects the information retrieval nature of the tasks (all SELECT operations). The \texttt{none} actions (2.9\%) represent cases where the model failed to produce a valid action, often indicating confusion or errors in understanding the current state.

\subsection{Quantitative Difficulty}
\label{subapp:task_difficulty}
For the generated enterprise web tasks, we introduce the \texttt{QuestionDifficultyAnalyzer} module to analyze task difficulty. 
Task difficulty is assessed from multiple dimensions, including Table complexity (number of tables and template types), Relational complexity (number of table relationships), Operation complexity (difficulty of \texttt{SELECT}/\texttt{INSERT}/\texttt{UPDATE}/\texttt{DELETE}), Result complexity (number of returned rows and columns) and SQL complexity (use of \texttt{JOIN}, \texttt{WHERE} clauses, subqueries, and aggregation operators). 
For example, the difficulty of a task is partly determined by the type of operation involved. 
Different operations are mapped to different difficulty values. 
We consider query operations (\texttt{SELECT}) to be relatively simple, while other operations such as creation (\texttt{INSERT}), update (\texttt{UPDATE}), and deletion (\texttt{DELETE}) are regarded as more complex due to their higher operational risk and execution complexity.

\subsection{Enviroment Infrastructure}
\label{subapp:enviroment_infrastructure}

For user-friendly use and reproducibility of our benchmark environment, 
we deploy each website in a separate Docker image inspired by \citep{zhou2023webarena}. 
The Docker images include all the code of the website, database, and any other software dependencies. 

They also do not rely on external volume mounts to function, as the data of the websites are also part of the docker image. 
This way, the image is easy to distribution containing all the pre-populated websites for reproducible evaluation. 
Our packaged Docker images can run users' systems and 
re-deploy the exact websites together with the data used in our benchmarks for their local benchmarking. 
For some tasks, the docker image may be reset to initial state after the evaluation.

\section{Details of Baseline Methods}
\label{app:mehods_details}

\subsection{Reinforcement Learning for EntAgent}
\label{subsec:rl}


Reinforcement learning (RL) \citep{paul2017rlhf,yao2023react,deepseek-r1} has shown promise in training large language models, but existing methods mainly focus on single-turn, non-interactive tasks. 
In contrast, real-world enterprise web agents operate in dynamic, interactive environments, where effective decision-making requires multi-turn reasoning and continuous information acquisition through interaction. 
In this work, we explore end-to-end policy optimization for training multimodal enterprise web agents that can actively interact with multi-turn environments and acquire richer information from sequential feedback via Group Relative Policy Optimization (GRPO) \citep{shao2024deepseekmath}. 
To facilitate stable learning, we first adopt supervised fine-tuning as a foundational stage. 
By fine-tuning the multimodal model on task trajectory data, the agent is equipped with essential prior knowledge of information systems and basic GUI interaction patterns, providing a strong and reliable initialization.
However, supervised fine-tuning primarily relies on static datasets that imitate user behaviors, which inherently limits the model’s adaptability to dynamic environments and its ability to handle long-horizon dependencies.
As a consequence, the web agent lacks self-evolution capabilities during multi-turn reasoning and often fails in high level tasks.
End-to-end multi-turn multimodal reinforcement learning naturally complements these limitations. 
Agent settings inherently require models to make sequential decisions, maintain memory across multiple turns, and adapt to stochastic environmental feedback \citep{yao2023react}. 
Based on supervised fine-tuning, the end-to-end training using reinforcement learning drives environment-grounded self-improvement, resulting in more robust and generalizable enterprise web agents.

\paragraph{Problem Formulation.}
In this section, we formulate the enterprise web task as a Markov Decision Process (MDP), defined by the tuple $(\mathcal{S},\mathcal{A},\mathcal{R})$. 
At each time step $t$, the GUI agent observes a state $s_t \in \mathcal{S}$ from the environment $E$, which is represented by the current web page's screenshot and textual content. 
Based on this observation, the agent predicts an action at from a predefined action space $\mathcal{A}$, which includes commonly used web operations. 
The GUI agent continues interacting with the environment in a multi-turn manner until the task is successful or the maximum step limit is reached. 
Throughout the interaction, the agent receives a scalar reward $r_t$ of the trajectory via the rule-based reward functions $\mathcal{R}$. 
Our objective is to maximize the rule-based reward from the environment over trajectories using GRPO. 

\paragraph{RL Algorithm.}
We adopt GRPO as our reinforcement learning algorithm to train the multimodal agents through end-to-end multi-turn interactions following the prior works \citep{qi2025webrl,wei2025webagent-r1}. 
GRPO eliminates the need for a value function by leveraging group-wise reward normalization to compute token-level advantages. 
This property makes it well-suited for training GUI agents with multiple image inputs and extended context length.

\paragraph{Reward Design.} 
To effectively guide policy optimization, we design a structured reward function incorporating both task-level success and action format correctness. 
To encourage more realistic and accurate actions during the interaction process, we assign rewards to step-level behaviors along the trajectory, which can be verified based on tool usage.

\begin{itemize}
    \item \noindent\textbf{Trajectory-Level Reward:} 
    For each task, we have a trajectory-level reward based on task completion. 
    A binary reward is assigned at the end of the trajectory: $\mathcal{R}_t = 1$ if the agent successfully completes the task, and $\mathcal{R}_t = 0$ otherwise. 
    This reward encourages the agent to learn effective multi-turn planning and execution.
    \item \noindent\textbf{Step-Level Reward:} 
    To mitigate error accumulation from intermediate steps and encourage realistic, accurate interactions, we introduce step-level behavioral rewards. 
    These rewards are verified through tool usage to ensure the executability of actions and environmental consistency. 
    if the action fails to execute successfully, we assign a penalty for this step. 
    The final step-level reward $\mathcal{R}_s$ is computed as the average over the trajectory, providing a fine-grained complement to the trajectory-level task reward.  
    \item \noindent\textbf{Format Reward:} 
    Each response generated by the GUI agent is parsed into a sequence of discrete actions. 
    We adopt the format rules from DeepSeek-R1 \citep{deepseek-r1} to ensure that the structured output format, where the model is required to place the reasoning within <think>...</think> and the final answer within </answer>...</answer>. 
    Responses that conform to the expected format receive a reward $\mathcal{R}_f = 1$; otherwise, $\mathcal{R}_f = 0$.
\end{itemize}
In both settings, we apply a rule-based reward function $\mathcal{R}$ composed of trajectory-level reward $\mathcal{R}_t$, step-level reward $\mathcal{R}_s$ and format reward $\mathcal{R}_f$:
\begin{equation}
\begin{aligned}
     \mathcal{R} = w_1\mathcal{R}_t + w_2\mathcal{R}_f + w_3\mathcal{R}_f
\end{aligned}
\label{fun.reward_final}
\end{equation}
where $w$ represents weight of each reward.

\paragraph{Trajectory Collection \& Training Pipeline.}
Training GUI agents with reinforcement learning requires scalable and efficient trajectory collection in rich interactive environments. 
To meet this requirement, we adopt an asynchronous rollout strategy designed for parallel interaction with real-time environments.
We build a pool of rollout workers, where each worker consists of an interactive environment paired with a GUI agent that maintains a history of screenshots and corresponding actions, represented as $((s_t, a_t))$. 
Each rollout worker continuously captures the current screenshot and textual information of the GUI environment and transmits it to a centralized language model inference server powered by vLLM \citep{kwon2023efficient}. 
The policy model processes these batched visual and textual observations in parallel and predicts the next action.
Parallel trajectory rollouts enable efficient utilization of GPU resources on the inference server and significantly reduce per-step decision latency.
Despite recent progress in variants of GRPO, training GUI agents remains challenging, particularly due to the sparsity of reward signals in complex web environments. 
As a result, rollouts often yield limited feedback, which can hinder effective policy optimization under GRPO.
To alleviate the reward sparsity caused by the need for exploratory interactions with the environment, we introduce an experience replay buffer that caches successful trajectories for each task. 
During training, if a GRPO training group consists entirely of failed trajectories (i.e., all rewards are zero), we randomly replace one trajectory with a previously stored successful trajectory from the corresponding task buffer. 
This mechanism ensures that once the agent successfully completes a task at least once, subsequent training groups will always include at least one rollout with a non-zero reward signal, thereby facilitating more stable and effective policy optimization.

\paragraph{EntAgent-RL Training}

Due to the scarcity of training tasks, a trajectory training set was specifically curated for this data using the the instruction of templates.
Based on the characteristics of the website data and task difficulty, these trajectories were divided into a trajectory fine-tuning set and a reinforcement learning training set according to website features and task complexity. 
The trajectory data were obtained through agent interactions based on different template instructions, and ultimately, trajectories that successfully completed the tasks were selected.


\subsection{Hyper-Parameter of the Baseline Agents}
We utilize GPT-4.1 \citep{achiam2023gpt}, Claude 3.5 Sonnet (via Computer Use API) \citep{anthropic2024computeruse} to evaluate the benchmark.  
We also employ UI-TARS-72B-DPO \citep{qin2025ui}, Qwen2.5-VL-32B-Instruct \citep{qwen2.5-vl}, WebRL \citep{qi2025webrl} and TTI \citep{shen2025tti}.
For these agents, we set the temperature parameter to 1.0, and top\_p to 0.9, and the maximum number of textual tokens for generation is set to 16384. 
We set the maximum steps of interaction to 30 and the maximum time limits to 20 minutes for all tasks since the agent could lead to a stuck environment under some unexpected cases.
For our EntAgent, we use Qwen2.5-VL-32B-Instruct model \citep{qwen2.5-vl} as the backbone model and train it using the trl framework \citep{vonwerra2022trl}. 
For trajectory rollout, we deploy 64 parallel virtual environments, with 8 rollouts collected per task, and conduct training for a single epoch. 
To encourage exploration, rollouts are generated with a temperature of 1.0. 
For policy optimization, we adopt the AdamW optimizer \citep{adamw} with a learning rate of 1e-6 and a mini-batch size of 8 per device, using a gradient accumulation step of 2. 
Following DAPO \cite{yu2025dapo}, we set the clipping parameters to $\epsilon_{low} = 0.2$ and $\epsilon_{high} = 0.3$ to balance exploration and exploitation. 

\subsection{The Prompts of The Baseline Agents}
\begin{figure}
    \centering
    \includegraphics[width=1\linewidth]{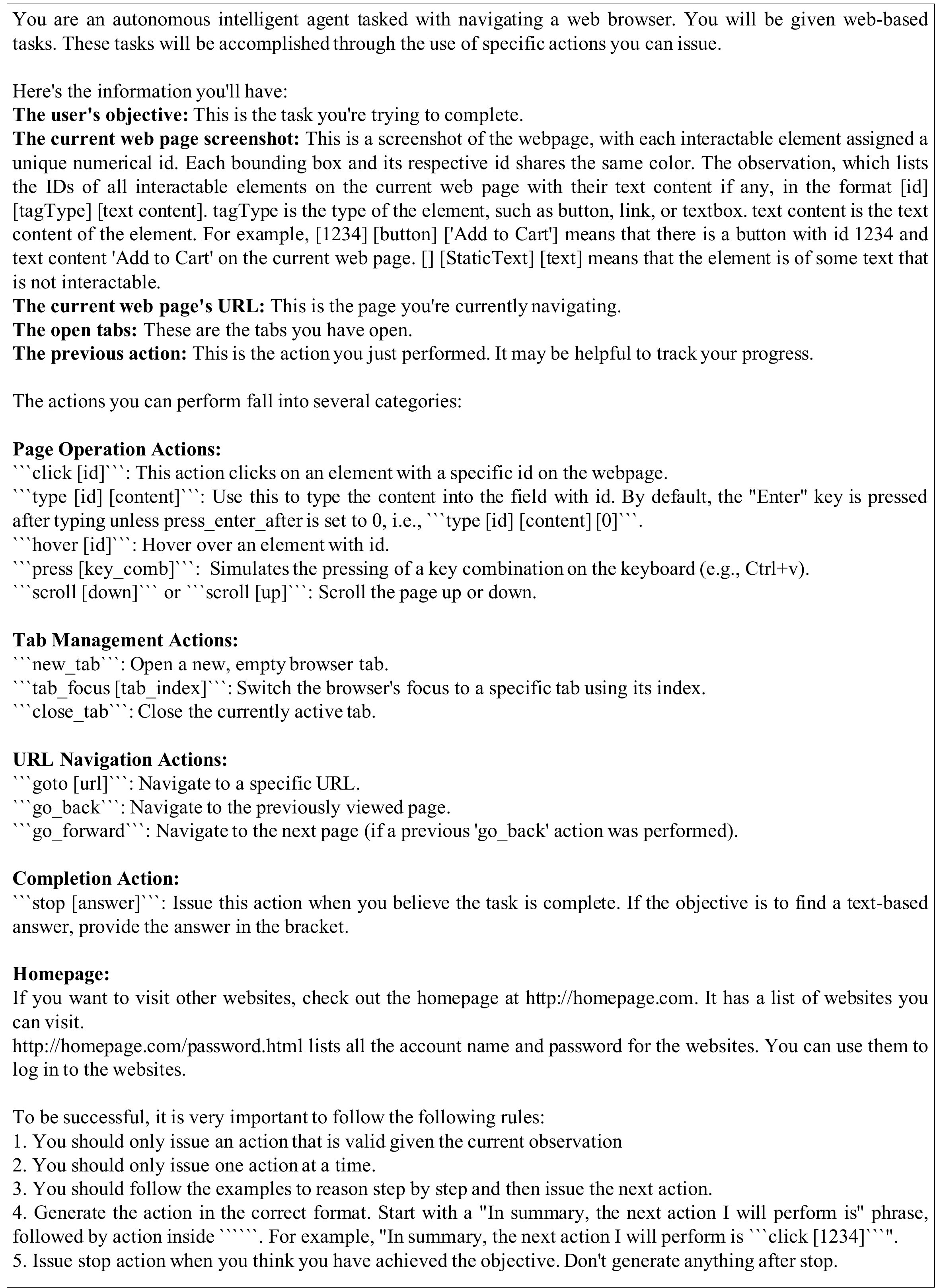}
    \caption{The system message of agents. This message includes the task object, the available actions and some notes.}
    \label{fig:prompt_system}
\end{figure}

To compare the performance of baselines, we use the sample system message of the reasoning agent for both GPT-3.5 and GPT-4 is in Figure \ref{fig:prompt_system}. 
In the system message, we have provided a brief explanation of the intents, observations, actions, etc., involved in GUI agent interactions, along with some important notes.

\section{Additional Case Studies}
\label{app:case_study}

\subsection{Cases based Agents Across Enterprise Business Domains}
\begin{figure}
    \centering
    \begin{minipage}{0.8\textwidth}
        \centering
        \includegraphics[width=1\linewidth]{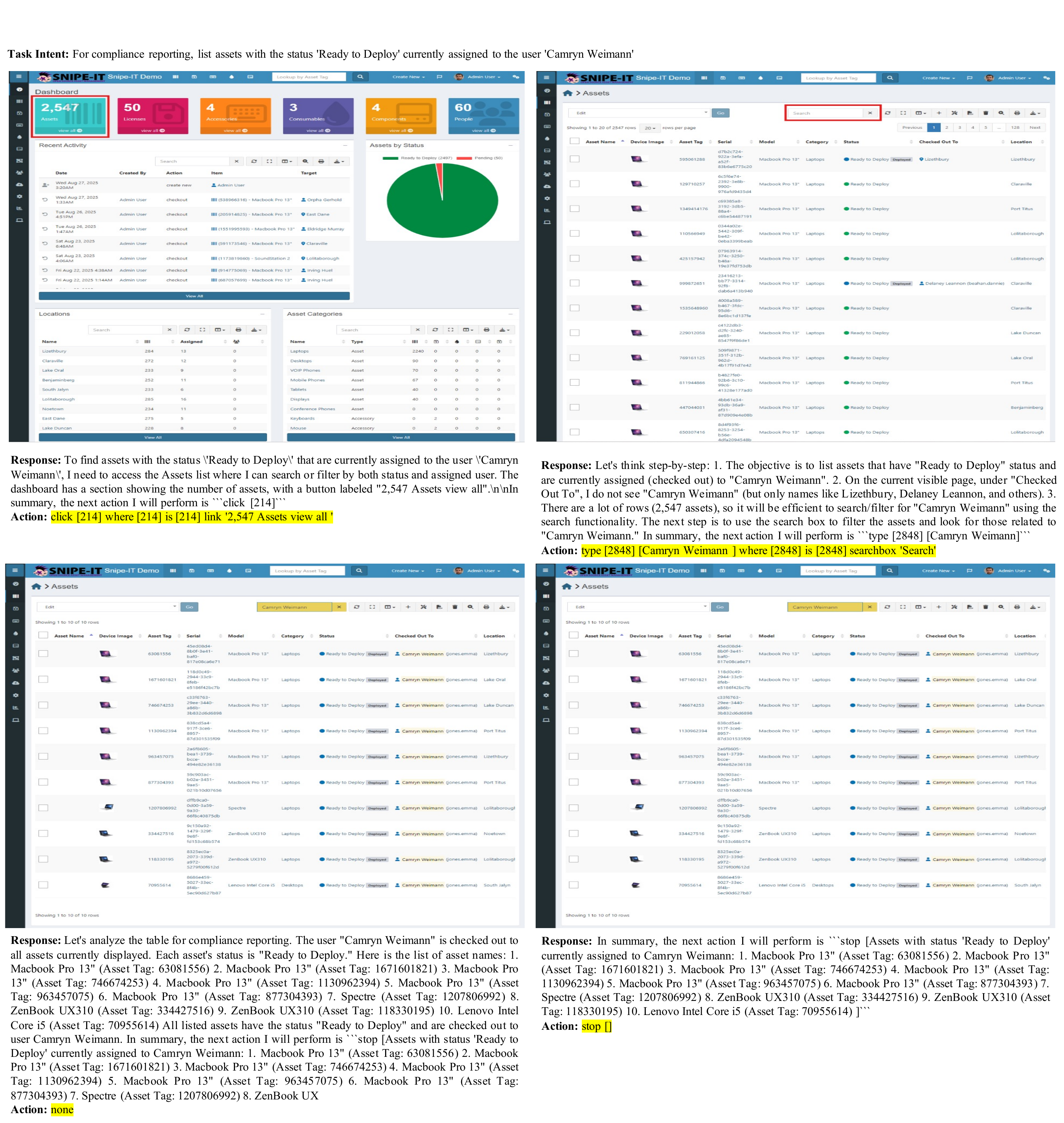}
        \caption{A case of agent, along with their screenshot and the accessibility tree of the relevant sections. The case about Snipe-IT is fail.}
        \label{fig:snipeit_80_failure}
    \end{minipage}
    \hfill
    \begin{minipage}{0.8\textwidth}
        \centering
        \includegraphics[width=1\linewidth]{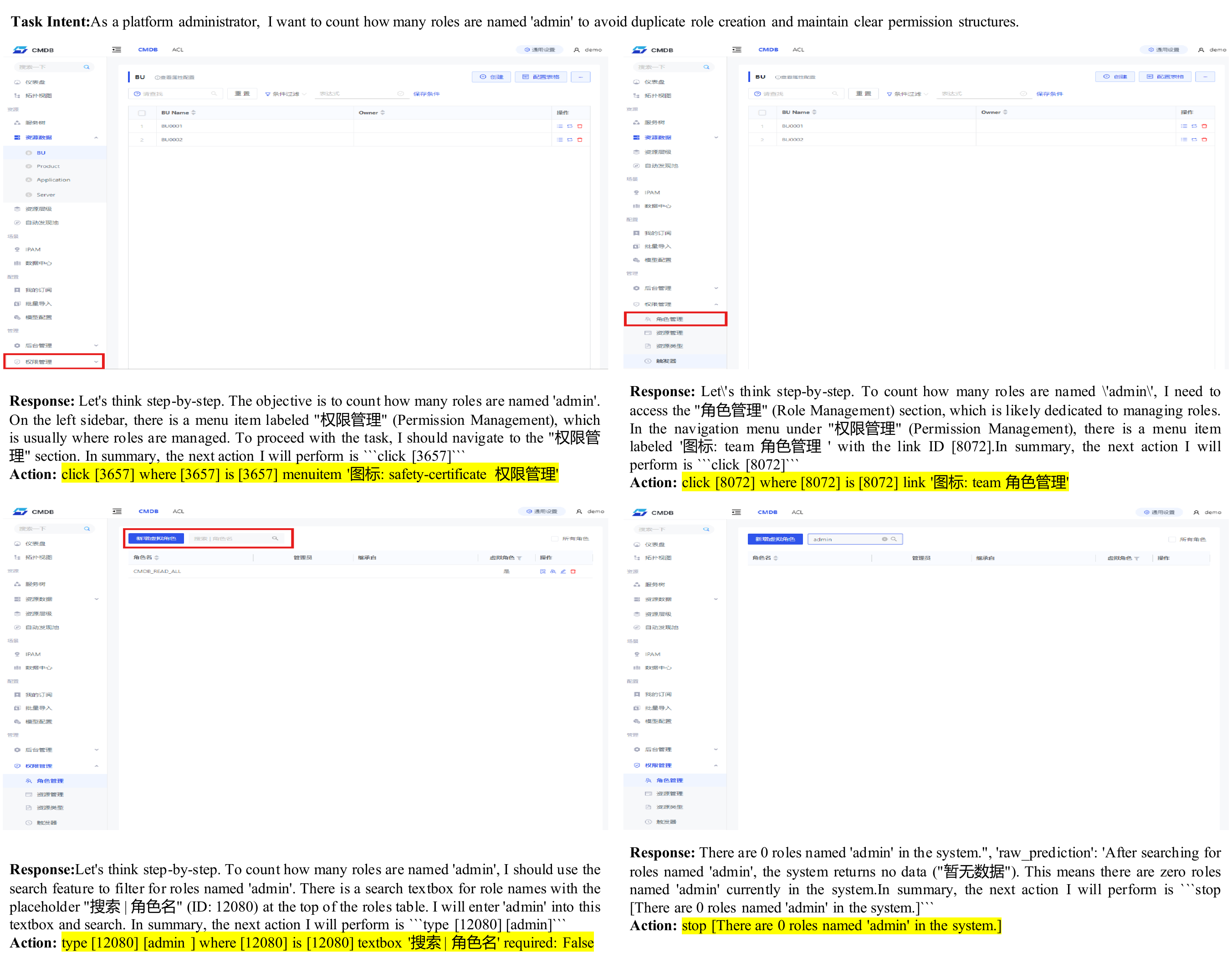}
        \caption{A case of agent, along with their screenshot and the accessibility tree of the relevant sections. The case about Veops CMDB is success.}
        \label{fig:veops_cmdb_18_failure}    
    \end{minipage}
\end{figure}

\begin{figure}
    \centering
    \begin{subfigure}{0.8\textwidth}
        \centering
        \includegraphics[width=1\linewidth]{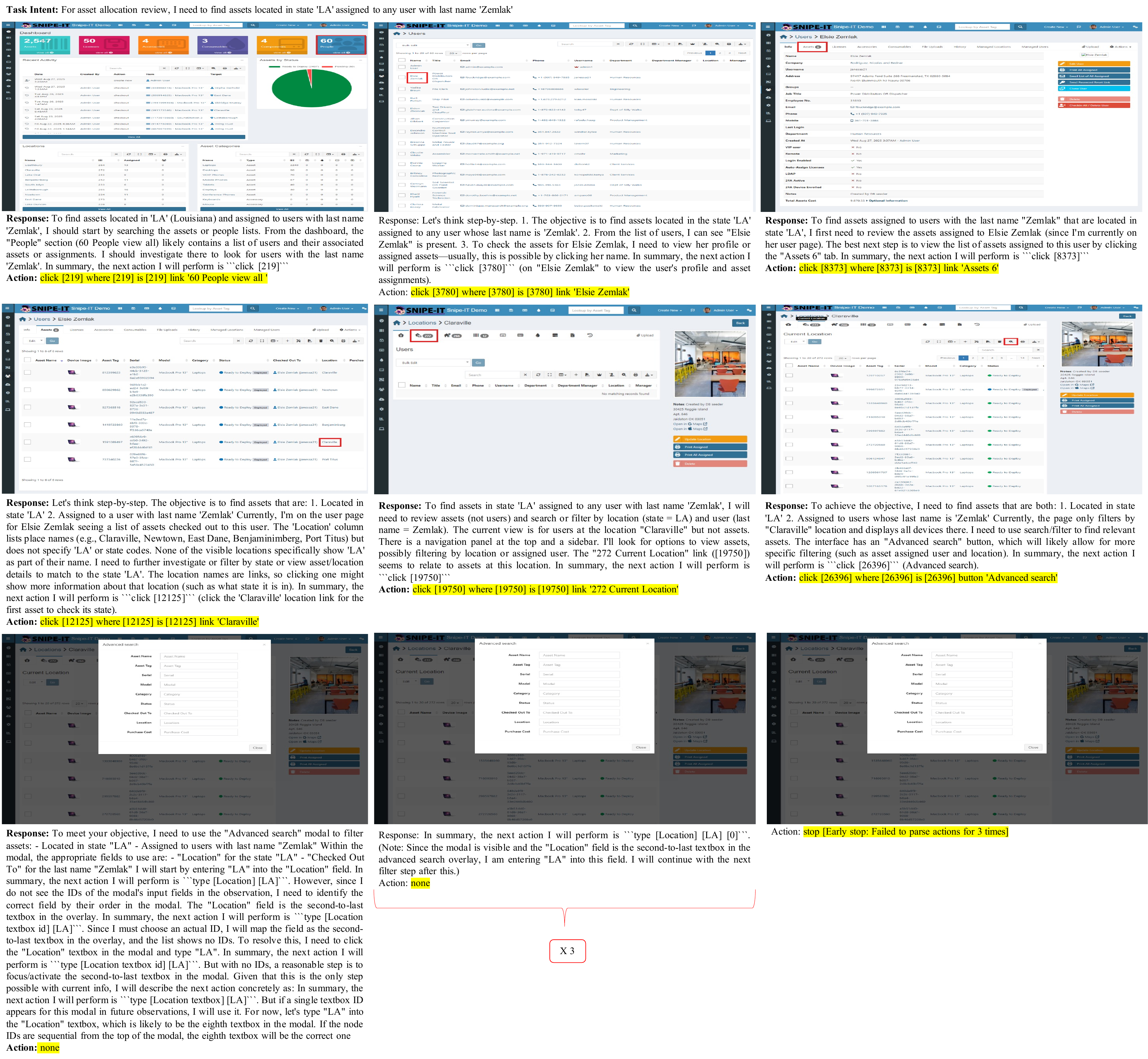}
        \caption{}
        \label{fig:snipit_83_gpt4.1}
    \end{subfigure}
    \hfill
    \begin{subfigure}{0.8\textwidth}
        \centering
        \includegraphics[width=1\linewidth]{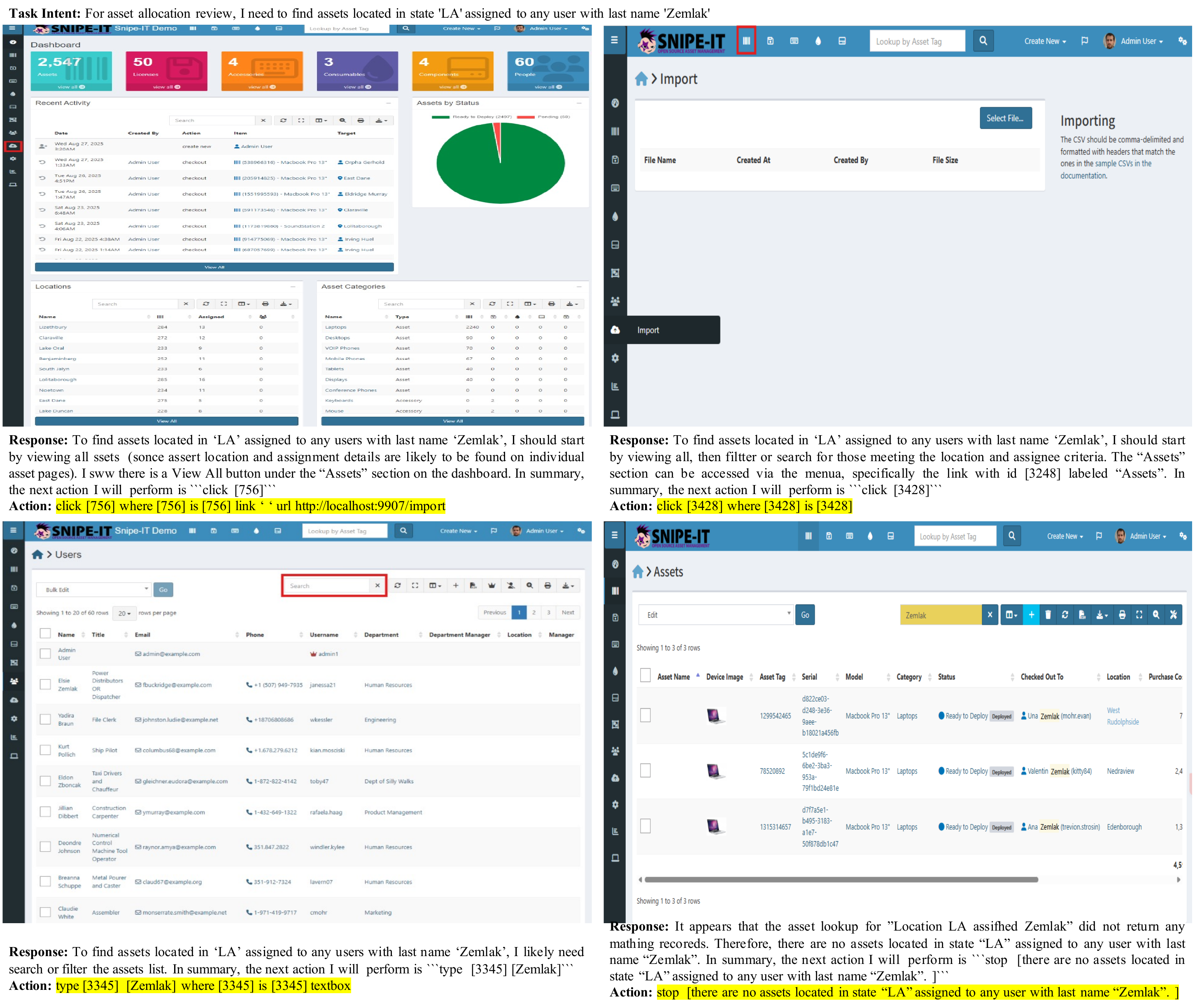}
        \caption{}
        \label{fig:snipit_83_rl}
    \end{subfigure}
    \caption{Comparison of GPT-4.1 and EntWorld-RL agents, along with their screenshot and the accessibility tree of the relevant sections. The case is about Snipe-IT. Figure \ref{fig:snipit_83_gpt4.1} is the result of GPT-4.1 agent. Figure \ref{fig:snipit_83_rl} is the result of EntWorld-RL agent.}
    \label{fig:comparison_case}
    
\end{figure}

Figures \ref{fig:snipeit_80_failure}–\ref{fig:comparison_case} illustrate the performance of different enterprise web tasks. 
Figure \ref{fig:snipeit_80_failure} is a failure case of GPT-4.1 agent. 
In the trajectory, the actions and observations look nice early, and the agent's subsequent responses are also quite reasonable. 
However, it might overlook certain other details, such as the required format for actions in the prompt. 
Even if the agent's response is semantically correct, errors in format can lead to parsing failures for actions, ultimately causing the task to fail. 
Figure \ref{fig:veops_cmdb_18_failure} is a success case of GPT-4.1 agent. 
In this example, it can be seen that the agent accurately understands the task intent and is able to make comprehensive judgments based on actions and page states to arrive at the correct answer. 
In Figure \ref{fig:comparison_case}, we compare the results of different agents in same enterprise web task. 
For GPT-4.1 Agent, it misunderstands the task intent, which lead to errors in the subsequent action sequence. 
Moreover, these errors accumulate over time. 
Later, it even triggers the limit on the maximum number of identical consecutive commands by repeatedly issuing the same instruction. 
As for EntWorld Agent, although deviations occur in its initial actions, it manages to adjust its strategy later. 
This adjustment helps steer it back on track, ultimately leading to a successful completion of the task.

\end{document}